\documentclass[lettersize,journal]{IEEEtran}
\usepackage{amsmath,amsfonts}
\usepackage{algorithmic}
\usepackage{algorithm}
\usepackage{array}
\usepackage[caption=false,font=normalsize,labelfont=sf,textfont=sf]{subfig}
\usepackage{textcomp}
\usepackage{stfloats}
\usepackage{url}
\usepackage{verbatim}
\usepackage{graphicx}
\usepackage{cite}
\usepackage{graphicx}
\usepackage{subfig}
\usepackage{svg}
\usepackage{caption}
\usepackage{subcaption}
\usepackage{amsmath}
\usepackage{booktabs}

\begin{document}

\title{A Hybrid Supervised and Self-Supervised Graph Neural Network for Edge-Centric Applications}

\author{Eugenio Borzone, \IEEEmembership{~Leandro Di Persia IEEE senior member} and Matias Gerard \thanks{
Eugenio Borzone, Leandro Di Persia and Matías Gerard are with Research institute for signals, systems and
computational intelligence (sinc(i)).

The authors would like to thank CONICET for the financial support provided throughout this research. We also extend our gratitude to the Research Institute for Signals, Systems, and Computational Intelligence (sinc(i)) in Santa Fe, Argentina, for their valuable resources and collaboration. Finally, a special thanks to ChatGPT for its assistance in the writing process.}}


\maketitle

\begin{abstract} 
This paper presents a novel graph-based deep learning model for tasks involving relations between two nodes (edge-centric tasks), where the focus lies on predicting relationships and interactions between pairs of nodes rather than node properties themselves. This model combines supervised and self-supervised learning, taking into account for the loss function the embeddings learned and patterns with and without ground truth. Additionally it incorporates an attention mechanism that leverages both node and edge features. The architecture, trained end-to-end, comprises two primary components: embedding generation and prediction. First, a graph neural network (GNN) transform raw node features into dense, low-dimensional embeddings, incorporating edge attributes. Then, a feedforward neural model processes the node embeddings to produce the final output. Experiments demonstrate that our model matches or exceeds existing methods for protein-protein interactions prediction and Gene Ontology (GO) terms prediction. The model also performs effectively with one-hot encoding for node features, providing a solution for the previously unsolved problem of predicting similarity between compounds with unknown structures.
\end{abstract}


\section{Introduction}
\IEEEPARstart{G}{raphs} are versatile structures used to model relationships between entities in non-Euclidean domains. They underpin applications ranging from social networks \cite{tiwari_social_nodate} and recommendation systems \cite{steck_deep_2021,covington_deep_2016} to bioinformatics \cite{wu2020comprehensive}.  
Early shallow embedding methods such as DeepWalk \cite{perozzi_deepwalk_2014}, LINE \cite{Tang_2015}, and Node2Vec \cite{grover_node2vec_2016} demonstrated the value of data-driven representations but share two well-known limitations: parameter redundancy and limited generalization \cite{Hamilton2017}.  
Graph Neural Networks (GNNs) overcome these drawbacks by parameter sharing and inductive learning \cite{zhou2020graph}.  Within GNNs, spectral and spatial formulations provide complementary views; the reader is referred to \cite{bruna2013spectral,defferrard2016convolutional,Shuman2013,ZhuRabbat2012,Bronstein2016,Wu2019} for canonical treatments.

\vspace{0.5em}
Message Passing Neural Networks (MPNNs) generalize earlier GNN variants by formalizing graph convolutions as iterative message-passing steps \cite{Gilmer2020}.  At iteration $t$, node $p$ aggregates information from its neighbours $\mathcal{N}(p)$ via
\begin{equation}
\begin{split}
m_{p}^{(t+1)} &= \sum_{q\in\mathcal{N}(p)}
                M_{t}\!\bigl(\mathbf{x}_{p}^{(t)},
                            \mathbf{x}_{q}^{(t)},e_{pq}\bigr),\\
\mathbf{x}_{p}^{(t+1)} &= U_{t}\!\bigl(\mathbf{x}_{p}^{(t)},
                          m_{p}^{(t+1)}\bigr),
\end{split}
\label{eq:generic_mpn}
\end{equation}

\noindent where $\mathbf{x}_{p}^{(t)}$ and $\mathbf{x}_{q}^{(t)}$ are the node features at step $t$, $e_{pq}$ is an optional edge feature, $M_{t}$ is the message function, $m_{p}^{(t+1)}$ is the aggregated message, and $U_{t}$ is the update function. After $T$ iterations, an order-invariant readout $R\!\bigl(\mathbf{x}_{p}^{(T)}\bigr)$ (e.g., sum or mean pooling) produces the final prediction at node or graph level. Equation~\eqref{eq:generic_mpn} establishes the notation used throughout the paper.

\subsection*{Motivation and contribution.}  
Existing MPNNs excel at node-level tasks but still under-utilize edge attributes and require large labelled datasets for supervision.  Edge-centric problems---typical in bioinformatics, where interactions are often encoded on edges \cite{wu2020comprehensive}---remain challenging. This work addresses these gaps by:

\begin{enumerate}
\item Proposing an MPNN-based architecture that natively treats node and edge features with a shared attention mechanism inspired by transformers \cite{Vaswani2017};
\item Introducing a hybrid loss that couples supervised and self-supervised objectives, enabling learning even when node features are sparse (e.g.\ one-hot encodings);
\item Demonstrating competitive performance on tasks that require edge-level regression and classification, including protein–protein interaction prediction.
\end{enumerate}

The remainder of this paper is organised as follows.
Section \ref{sec:materials_methods} details the materials and methods that ground the proposed model; Section \ref{sec:datasets} introduces the benchmark datasets employed in our experiments; Section \ref{sec:results} reports and discusses the empirical results and ablation studies; finally, Section \ref{sec:conclusion} summarises the main contributions and outlines future research directions.

\section{Materials and methods}
\label{sec:materials_methods}
This work presents a neural model composed of two main parts: an embedding block and a prediction block. The prediction block produces an estimation for pairs of nodes in the graph. This section presents the theoretical concepts and materials used in this work. First, we describe the embedding block. Then, we show the prediction block and its components, and explain the process of building the dataset.

\subsection{Graph and subgraph definitions}
\begin{figure}[htb]
    \centering
    \subfloat[]{
        \includegraphics[width=0.2\textwidth]{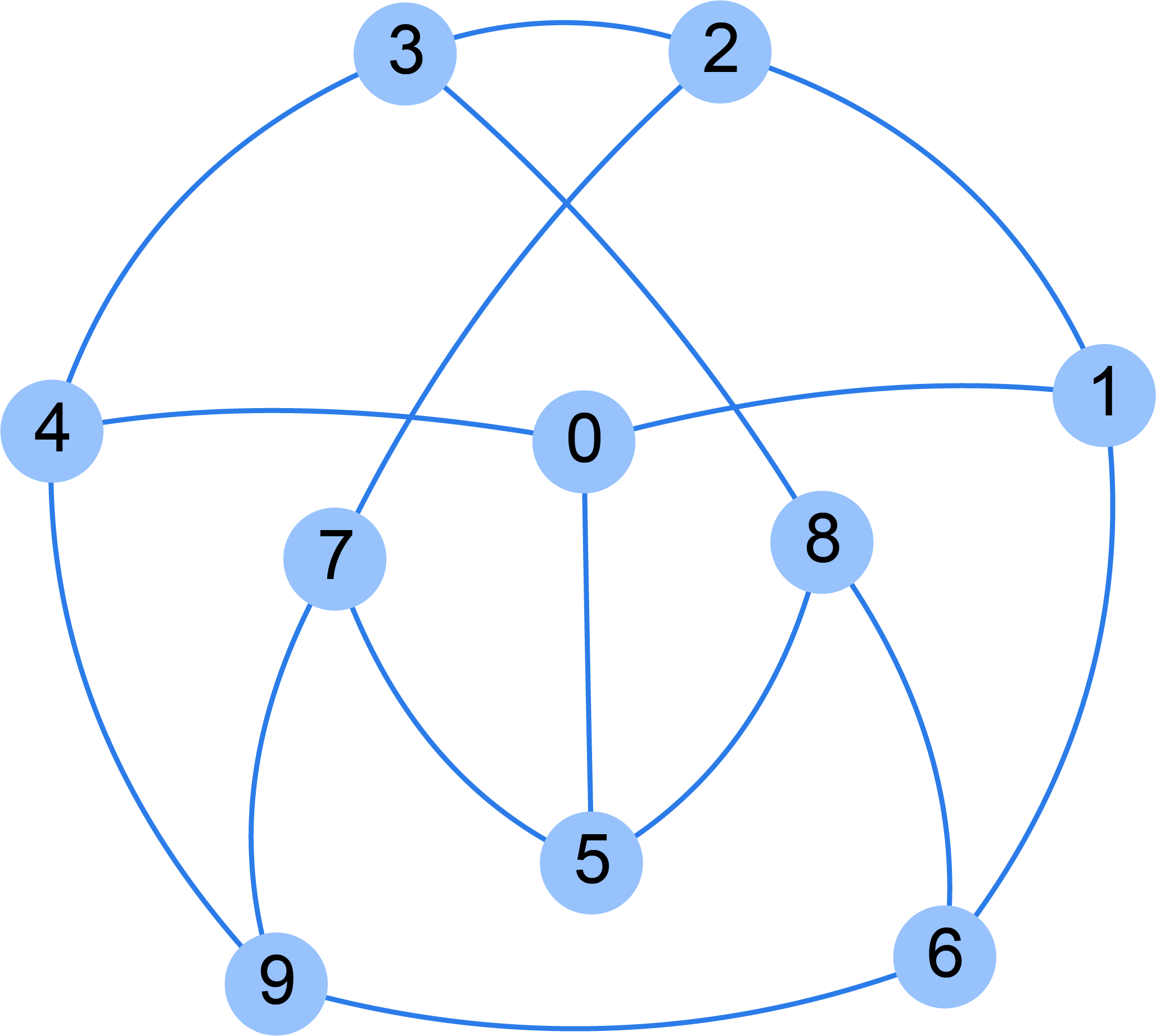}
    }
    \hfill
    \subfloat[]{
        \includegraphics[width=0.2\textwidth]{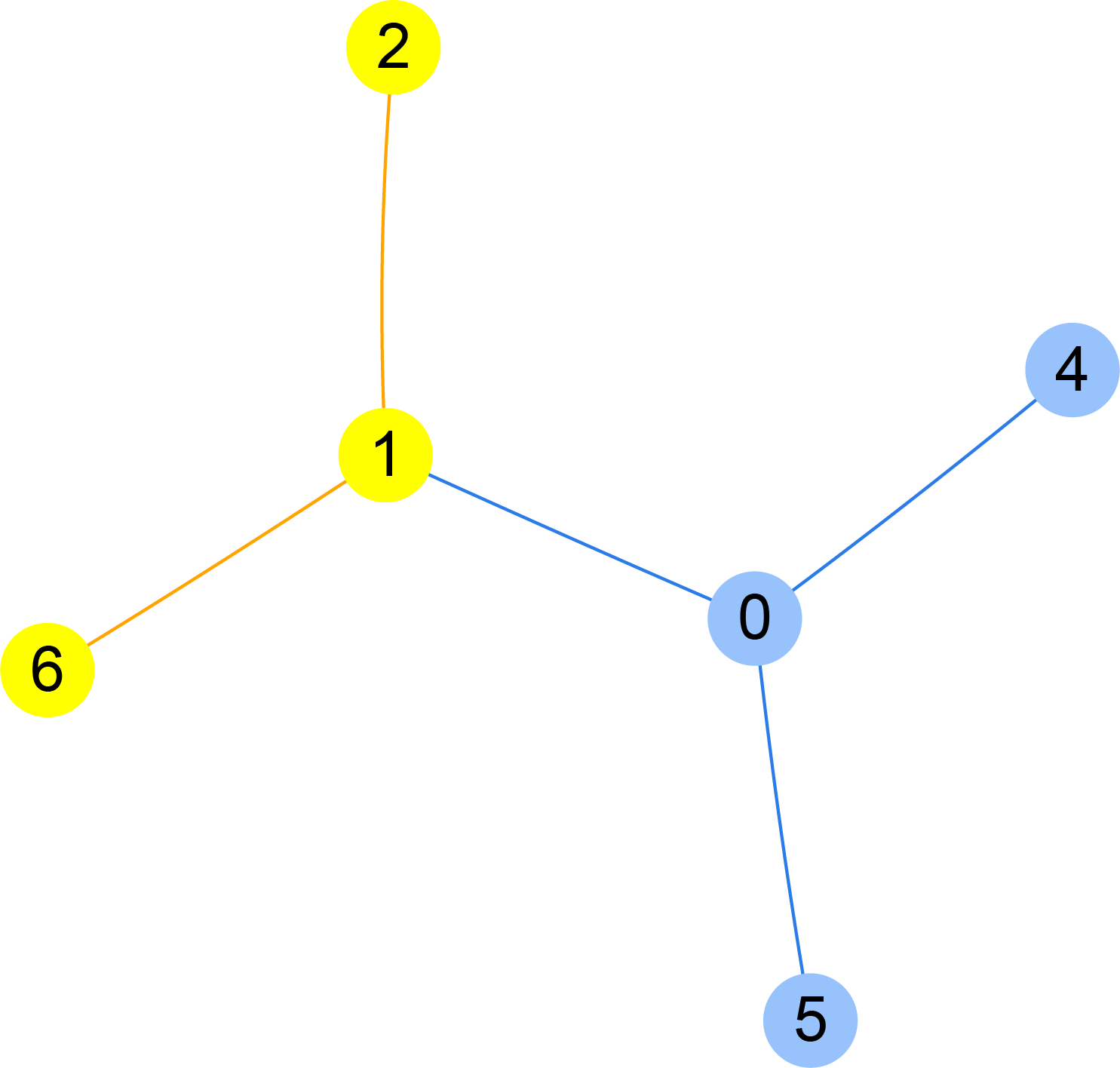}
    }
    \caption{(a) Example of a graph $\mathcal{G}$. (b) Example of pattern subgraph $\mathcal{G}'_{0,1}$ induced for nodes 0 and 1. Nodes and edges related to node 1 are highlighted in yellow, while those related to node 0 are shown in blue. These represent the first-degree neighbors of each node within the subgraph. These graphs represent only the structure; in this representation, each node is associated with a feature vector, and each edge has corresponding edge features.}
    \label{fig:network-pattern}
\end{figure}

Let $\mathcal{G} = \{v, e\}$ be a graph, where $v$ is the set of $|v|$ nodes in the network, and $e$ denotes the set of edges connecting them. Let $\mathbf{X} \in \mathbb{R}^{|v| \times d}$ be the feature matrix of nodes in $\mathcal{G}$, where $d$ is the number of features for each node. Furthermore, let $\mathcal{E} \in \mathbb{R}^{|e| \times q}$ denote the edge feature matrix, where each row represents the features associated with a specific edge in the graph. These edge features may or may not exist and can be a vector of arbitrary size $q$. Figure \ref{fig:network-pattern}a shows an example of a typical graph and its elements.

The subgraph $\mathcal{G}'_{i,j} = \{v_{i,j}, e_{i,j}\} \subseteq \mathcal{G}$ is defined as being induced by nodes $i$ and $j$ (Figure \ref{fig:network-pattern}b), where $v_{i,j} = \{\mathcal{N}(i), \mathcal{N}(j)\}$ are the nodes included, and $e_{i,j}$ are the connections between them. The set $\mathcal{N}(k)$ is represented by the direct neighbors that share an edge with node $k$. Additionally, the edge features matrix $\mathbf{E}'_{\mathcal{G}'_{i,j}} \in \mathbb{R}^{ |e_{i,j}| \times q}$ is associated with $|e_{i,j}|$ number of connections in the subgraph $\mathcal{G}'_{i,j}$. The feature matrix $\mathbf{X}'_{\mathcal{G}'_{i,j}} \in \mathbb{R}^{|v_{i,j}| \times d}$ is also defined for the $|v_{i,j}|$ nodes in the subgraph $\mathcal{G}'_{i,j}$.

\subsection{Proposed model}
Our model operates on subgraphs $\mathcal{G}'$, structured around two central nodes. This approach is essential for edge-centric tasks, where the aim is to predict the presence or properties of connections between specific pairs of nodes. The subgraph is constructed to enrich the representation of these central nodes by leveraging their local neighborhood. In each subgraph, the central nodes and their direct neighbors, along with all existing edges among these nodes, form the context for learning.

\begin{figure*}[!t]
\centering
\includegraphics[width=\textwidth]{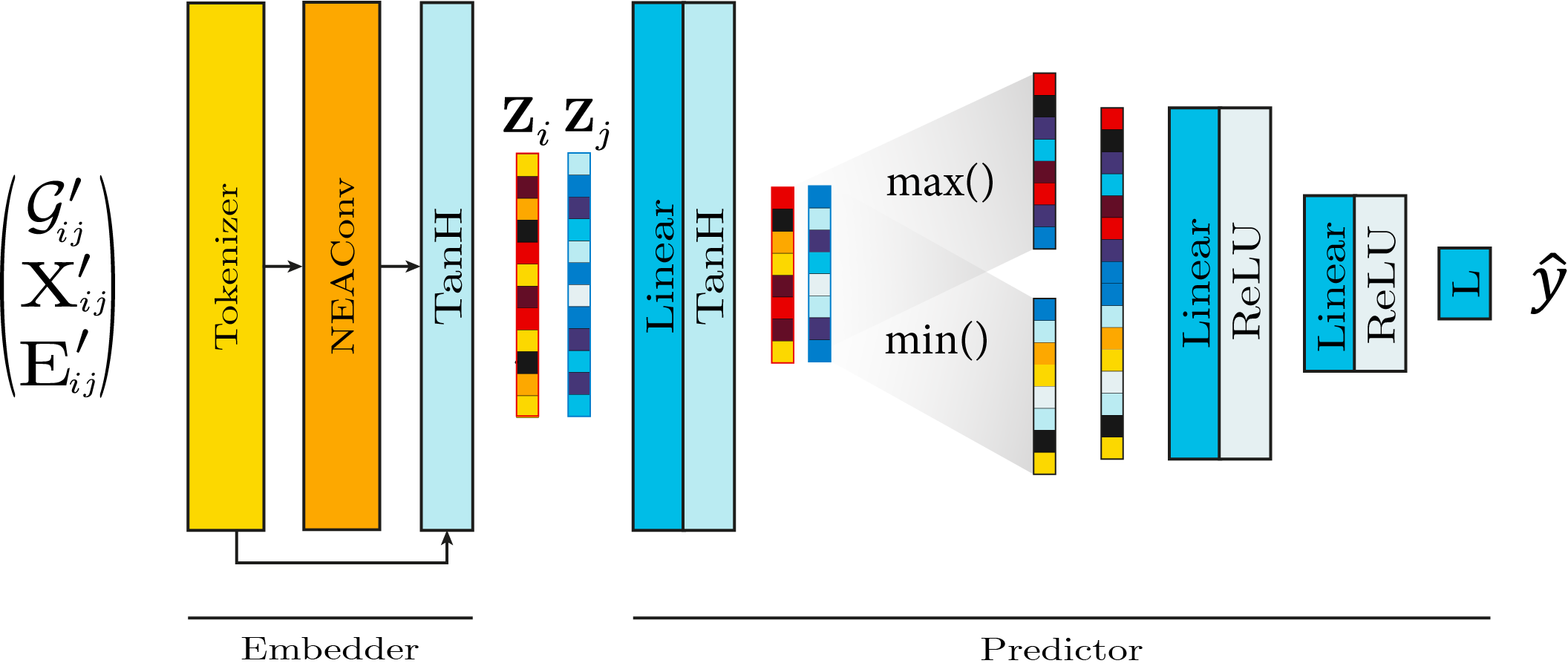}
\caption{Model architecture overview. The input consists of the pattern subgraph $\mathcal{G}'_{i,j}$, node features $\mathbf{X}'_{\mathcal{G}'_{i,j}}$, and edge features $\mathbf{E}'_{\mathcal{G}'_{i,j}}$. These inputs are processed through the embedding generation block: Tokenizer, the \emph{NodeEdgeAttentionConv} (\emph{NEAConv}) layer, and Tanh to form the embeddings $Z_i$ and $Z_j$. The embeddings then pass through a linear layer followed by a Tanh activation, are reordered, and finally fed into a three-layer perceptron to produce the output.}
\label{fig:model}
\end{figure*}

The model processes each subgraph in two stages: embedding generation and prediction. In the embedding generation phase, the node features within the subgraph are transformed into dense, low-dimensional vectors $\mathbf{Z}$ that capture both structural and feature-based information. These embeddings, particularly those of the central nodes, are then concatenated and passed to the prediction module to estimate the desired property associated with their connection. The entire model is trained end-to-end using a loss function that combines supervised and self-supervised training, allowing it to learn meaningful patterns even for nodes with limited or missing initial information. Figure \ref{fig:model}  illustrates the model's architecture, with detailed descriptions of each component provided in the following sections.

\subsection{Embedding generation}
This block involves two components: the tokenizer and the \emph{NodeEdgeAttentionConv} (\emph{NEAConv}) layer, as illustrated in Figure \ref{fig:model}. Node embeddings are compact, low-dimensional representations of graph nodes, integrating both node and edge features. Our model effectively generates these embeddings, capturing the underlying graph structure and node characteristics even when one-hot encoding is used as node features. This capability ensures efficient information propagation and accurate downstream predictions. We focus on the architecture and mechanisms used in our model to produce these meaningful representations that encode the underlying graph structure, node characteristics, and edge information.

The tokenizer takes the node features $\mathbf{X}'_{\mathcal{G}'_{i,j}}$ as input and projects them into a more suitable space for inference using a Multi-Layer Perceptron (MLP), yielding a new representation $\widetilde{\mathbf{X}}'_{\mathcal{G}'_{i,j}}$. This representation is then processed by a single \emph{NEAConv} layer, a message-passing neural network specifically designed for edge-centric tasks. In this layer, custom message and update functions are defined to align with our model architecture.

\begin{figure}[!t]
\centering
\includegraphics[width=0.4\textwidth]{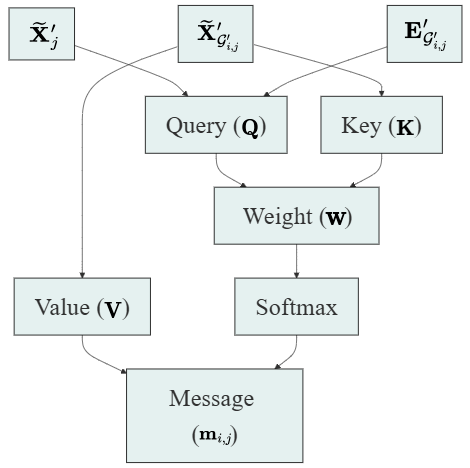}
\caption{Diagram illustrating the attention mechanism in the message function. It shows how node and edge features are transformed into key, query, and value vectors to assign attention weights, facilitating effective information aggregation from neighboring nodes.}
\label{fig:message}
\vspace{-10pt}
\end{figure}

Figure \ref{fig:message}, shows the message-passing process involved in the \emph{NEAConv} layer. The information of each target node $i,j \in \mathcal{G}'_{i,j}$ is updated through the attention mechanism described in Figure \ref{fig:message}. For a given node $j$, the algorithm initially projects the tokenized features of the node $\widetilde{\mathbf{X}}'_{j}$ and those of all the pattern $\widetilde{\mathbf{X}}'_{\mathcal{G}'_{i,j}}$ into three new spaces with the same dimension as the original.  

These nonlinear projections are performed by three separate perceptrons, each consisting of a linear layer followed by a sigmoid activation. Because the propagated information comes exclusively from first-order neighbours rather than from an ordered sequence, the incoming messages are intrinsically unsorted; therefore, no positional encoding is applied. The query vector $\mathbf{Q}$ is generated from the concatenation of the target node features $\widetilde{\mathbf{X}}'_{j}$ and, when available, the edge features $\mathbf{E}'_{\mathcal{G}'_{i,j}}$, capturing the relevance of the target node within the context of its surrounding neighborhood. The key vector $\mathbf{K}$ is derived from the source nodes' features $\widetilde{\mathbf{X}}'_{\mathcal{G}'_{i,j}}$, encoding essential information about the source nodes to be compared with the query vector. The value vector $\mathbf{V}$, also generated from the source nodes' features $\widetilde{\mathbf{X}}'_{\mathcal{G}'_{i,j}}$, carries the information that will be propagated to the target node.  The vector weight $\mathbf{w}$ which represents the importance of each source node's information for the target node, is calculated as:

\begin{equation}
\mathbf{w} = \text{sum}(\mathbf{Q} \odot \mathbf{K}, \text{dim}=1)
\end{equation}
where $\mathbf{Q} \odot \mathbf{K}$ represents the element-wise (Hadamard) product of the two tensors. This vector $\mathbf{w}$ is then normalized using a \textit{softmax} function. The normalized vector $\mathbf{w}$ is then used to construct the new representation $\widehat{\mathbf{X}}'_{j}$ of the central node, through the weighted combination of the representations $\mathbf{V}$ of the neighboring nodes. The \textit{update function} concatenates the aggregated messages: 

\begin{equation}
\widehat{\mathbf{X}}'_{j} = \sum_{k \in \mathcal{N}(j)} \
\mathbf{m}_{k,j}
\end{equation}
with the original tokenized node features $\widetilde{\mathbf{X}}'_{j}$ resulting in updated embeddings that integrate both the aggregated information and the original features: 

\begin{equation}
{\mathbf{Z}_{j} = Tanh \left ( \left [ \widehat{\mathbf{X}}'_{j}, \widetilde{\mathbf{X}}'_{j} \right ] \right )}
\end{equation}

\subsection{Prediction step}
The prediction module in our model is a multilayer perceptron (MLP) designed to process the generated node embeddings and produce task-specific predictions. To ensure that the model predictions are invariant to the permutation of the nodes, we first perform a reordering of the features from the node embeddings. This reordering helps in capturing the important features across different nodes while maintaining the permutation invariance property.
Initially, we extract the minimum and maximum values along the feature dimension of the node embeddings. These extracted features are then concatenated to form a new feature representation that encapsulates the critical information from the node embeddings.
The architecture of the MLP consists of three linear layers, each followed by an activation function, to produce the final prediction output.

\subsection{Loss function}
The proposed loss function is build for optimizing node embeddings and predictions simultaneously. This is defined as:
\begin{equation}
\mathcal{L} = \alpha \mathcal{L}_{\text{supervised}} + \beta \mathcal{L}_{\text{cosine}} + \gamma \mathcal{L}_{\text{cosine\_pred}}
\label{eq:loss}
\end{equation}
where $\alpha$, $\beta$, and $\gamma$ denote weighting coefficients.
Preliminary experiments showed that the three loss components are of the same order of magnitude; accordingly, we fixed $\alpha=\beta=\gamma=1$ to avoid privileging any single term.
These coefficients may, however, be treated as hyper-parameters and tuned for particular datasets in future work.

$\mathcal{L}_{\text{supervised}}$ and $\mathcal{L}_{\text{cosine}}$ are supervised terms that guide the model to align its predictions and embeddings with the true labels, while $\mathcal{L}_{\text{cosine\_pred}}$ serves as a self-supervised term, refining the embeddings by aligning the predictions with the cosine similarity of the node embeddings. All three terms follow the same structure: for regression tasks, the Mean Squared Error (MSE) is used, and for classification tasks, the cross-entropy loss is applied.

The term $\mathcal{L}_{\text{supervised}}$ minimizes the difference between the predicted outputs $\hat{Y}$ and the true labels $Y$, ensuring accurate predictions. Similarly, $\mathcal{L}_{\text{cosine}}$ compares the cosine similarity of the node embeddings, denoted as $\widetilde{Y}$ (see equation \ref{eq:wideY}), to the ground truth $Y$, guiding the model to learn embeddings that reflect meaningful relationships between nodes. Both terms apply MSE for regression and cross-entropy for classification, depending on the task.

\begin{equation}
    \widetilde{Y} = cosine\_similarity(\mathbf{Z}_i,\mathbf{Z}_j)
    \label{eq:wideY}
\end{equation}

In addition to the supervised terms, the loss function includes a self-supervised component, $\mathcal{L}_{\text{cosine\_pred}}$, which aligns the predicted output $\hat{Y}$ with the cosine similarity of the node embeddings $\widetilde{Y}$. This self-supervised term helps the model learn robust representations even for nodes with unknown or incomplete information by structuring the embeddings in a meaningful way. As a result, the model becomes more adaptable to scenarios where data is missing, improving its generalization and performance. During training, if an input pattern does not have true labels assigned, only the third term will be used, allowing the model to learn in a unsupervised way.

The combination of these terms allows the model to learn more robust and generalizable embeddings, as well as make accurate predictions. The inclusion of nodes with unknown information in the loss function ensures that the embeddings capture the overall structure and relationships within the graph, improving the representation of all nodes, regardless of the availability of explicit labels.

\section{Datasets}
\label{sec:datasets}

Experimental validation covers three widely used bioinformatic tasks in the literature: \emph{Protein–Protein Interaction} (PPI) prediction, \emph{Gene Ontology} (GO) annotation, and \emph{metabolic-compound similarity}. 

\subsection{Protein–Protein Interaction}
\label{sec:ppi}
Five balanced PPI benchmarks introduced by Yang \textit{et al.}~\cite{Yang2020}\footnote{\url{http://www.csbio.sjtu.edu.cn/bioinf/LR_PPI/Data.htm}} are employed: \textit{HPRD}, \textit{Human}, \textit{E.\ coli}, \textit{Drosophila} and \textit{C.\ elegans}.  
For every dataset a $k$-nearest-neighbour graph (KNNG) is built from ClustalO sequence similarities \cite{Sievers2011,Paredes2006}; $k$ is tuned on the validation split.  
Node features use Composition-of-Triads (343-dim.) representations \cite{Shen2007}, and pairwise similarity scores act as edge attributes.

\subsection{Gene Ontology Terms}
\label{sec:go}
Following the EXP2GO protocol \cite{exp2go}, GO annotation transfer is evaluated on \textit{Saccharomyces cerevisiae}, \textit{Arabidopsis thaliana} and \textit{Dictyostelium discoideum}.  
Expression profiles (79–740 measurements per gene) from ~\cite{Eisen1998,Espinoza2010,Kreppel2004} constitute node features, while semantic similarities define weighted edges.  Annotations are filtered as in CAFA \cite{Zhou2019}.  The KNNG construction and $k$ selection mirror the PPI pipeline.

\subsection{Metabolic Pathways}
\label{sec:metabolic}

Expanding our previous glycolysis study \cite{Borzone2022b,Borzone2022a}, the benchmark now encompasses six KEGG pathways: \textit{Glycolysis}, \textit{Starch and sucrose metabolism}, \textit{Pentose phosphate pathway}, \textit{Citrate cycle (TCA cycle)}, \textit{Pyruvate metabolism}, and \textit{Propanoate metabolism}.  
The resulting graph contains 207 compounds, of which 174 have known SMILES structures and 33 lack structural information (\emph{unknown compounds}).  MACCS fingerprints \cite{Durant2002} computed with RDKit\footnote{\url{https://www.rdkit.org}} serve as node features for the known compounds; one-hot placeholders mark unknown structures.  Edge weights are Tanimoto coefficients \cite{Bajusz2015}, and enzyme-family one-hot vectors (EC 1–7) supply additional edge attributes.  In total, 21\,528 compound pairs are generated, including 5\,390 pairs that involve at least one unknown compound.

\section{Results}
\label{sec:results}
The performance measure for protein-protein interaction (PPI) prediction was evaluated using the F1-score, which is derived from precision and recall. Precision is defined as the ratio of true positive predictions to the total predicted positives, expressed mathematically as:

\begin{equation}
\text{Precision} = \frac{TP}{TP + FP}
\end{equation}
where \( TP \) represents the number of true positives and \( FP \) represents the number of false positives. 

Recall, also known as sensitivity, is defined as the ratio of true positive predictions to the total actual positives, given by the equation:

\begin{equation}
\text{Recall} = \frac{TP}{TP + FN}
\end{equation}
where \( FN \) denotes the number of false negatives. 

The F1-score is the harmonic mean of precision and recall, providing a single metric that balances both measures. It is calculated using the following equation:

\begin{equation}
F1 = 2 \times \frac{\text{Precision} \times \text{Recall}}{\text{Precision} + \text{Recall}}
\end{equation}

This score is particularly useful in the context of PPI prediction, as it takes into account both the accuracy of positive predictions and the ability to identify all relevant interactions.

For GO terms prediction results are reported according to the CAFA rules \cite{Zhou2019}, with the maximum F1-measure ($\text{F}1_{max}$), which considers predictions across the full spectrum from high to low sensitivity. The calculation of $\text{F}1_{max}$ is conducted through a systematic approach to evaluate the quality of predictions in biological functions. First, the results of the predictions and the actual annotations for the functions being assessed are collected. Next, precision and recall are calculated for each function, considering all possible thresholds to observe how these metrics vary. The decision thresholds are then adjusted, and for each threshold, the F1 measure, which combines precision and recall into a single metric, is computed. $\text{F}1_{max}$ is defined as the maximum F1 value obtained by varying the thresholds, allowing for the identification of the threshold that maximizes this metric. Ultimately, a higher $\text{F}1_{max}$ value indicates a better balance between precision and recall, reflecting greater effectiveness in the predictions made.

For similarity prediction  between compounds, the Mean Absolute Error (MAE) was utilized to report performance. The MAE is defined as the average of the absolute differences between predicted and actual values, providing a measure of the accuracy of predictions. It is calculated using the following equation:

\begin{equation}
\text{MAE} = \frac{1}{n} \sum_{i=1}^{n} |y_i - \hat{y}_i|
\end{equation}
where \( n \) represents the number of predictions, \( y_i \) denotes the actual value, and \( \hat{y}_i \) indicates the predicted value. The use of MAE allows for a straightforward interpretation of prediction errors, as it reflects the average magnitude of errors in a set of predictions without considering their direction.
\begin{figure*}[t]
\centering
\includegraphics[width=\textwidth]{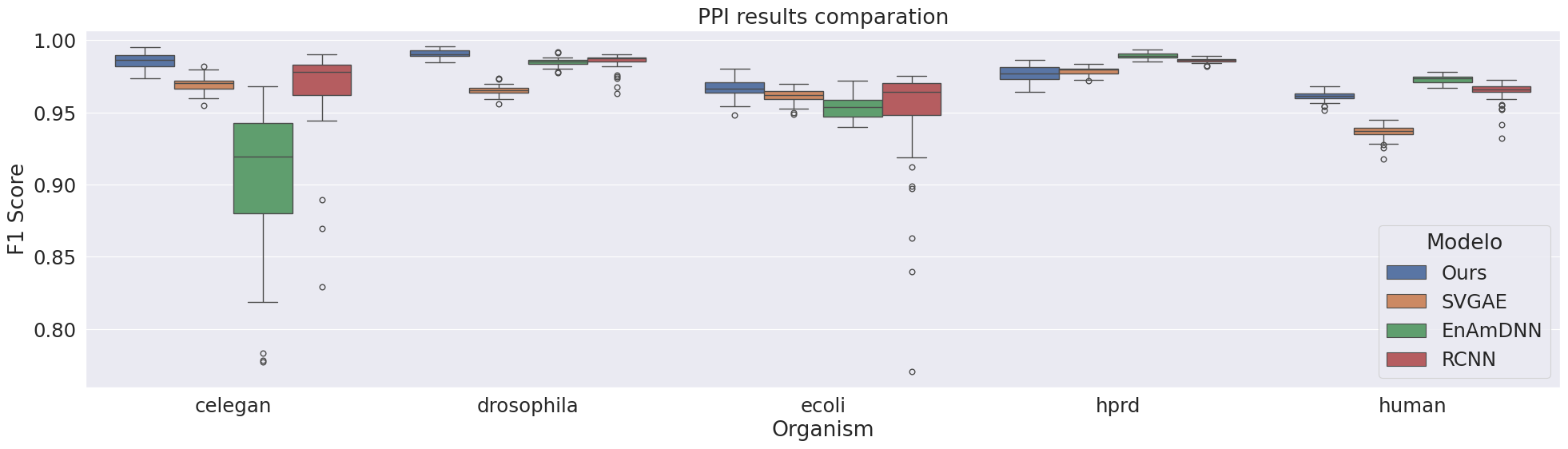}
\caption{Comparison of our model with SVGAE, Siamese Residual RCNN, and EnAmDNN on five PPI organisms. Each boxplot represents the distribution of the F1 scores obtained from the 10 times 5-fold cross-validation. The boxplots show the mean F1 score for each k-fold.}
\label{fig:ppi_boxplot}
\end{figure*}

\subsection{Protein-Protein Interaction (PPI) prediction}
The experiment was conducted using a 5-fold cross-validation, repeated 10 times, to mitigate the effect of random initialization. This procedure averaged the results over multiple runs, reducing variability in the final performance metrics. The same folds were used by our model and the baseline for a fair comparison. Preliminary experiments indicate that an input tokenizer composed of two linear layers using ReLU and a final Tanh activation function produced the best results.

\begin{figure}[h]
    \centering
    \includegraphics[width=\linewidth]{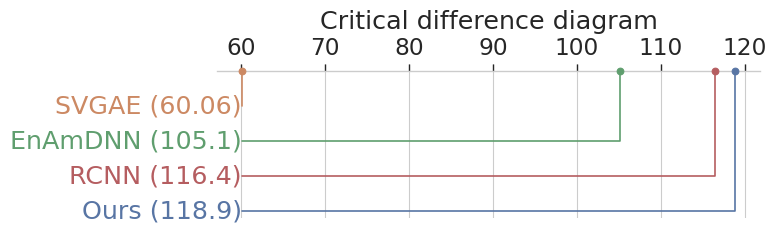}
    \caption{Critical-difference (CD) diagram of average ranks for the PPI experiments. With a CD value of 1.83 ($\alpha = 0.05$), no horizontal black lines appear, meaning every pair of models differs by more than the CD and thus all performance differences are statistically significant.}
    \label{fig:critical_ppi}
\end{figure}

To evaluate the performance of the proposed model, we compared it against several state-of-the-art approaches. First, we included the Signed Variational Graph Auto-Encoder (SVGAE) \cite{Yang2020}, a method specifically designed for PPI tasks that uses graph structures to capture complex relationships in biological networks, achieving superior performance compared to traditional sequence-based models. Additionally, we considered the Siamese Residual RCNN model proposed by Chen et al. \cite{Chen2019}, a deep learning approach that applies residual connections and convolutional architectures to predict PPIs from sequence information. Finally, we evaluated EnAmDNN (Ensemble Deep Neural Networks with Attention Mechanism) \cite{Li2020}, the leading sequence-based predictor reported in the recent benchmark by Dunham et al. \cite{Dunham2022}, which demonstrated remarkable accuracy by integrating ensemble learning and attention mechanisms.

Figure~\ref{fig:ppi_boxplot} summarises the F1-score distributions for the PPI task.  
For each of the five organisms, four side-by-side boxplots compare our model with SVGAE, Siamese Residual RCNN, and EnAmDNN. Every boxplot reflects the scores obtained in 10 repetitions of 5-fold cross-validation; the line inside the box denotes the mean F1 of each k-fold.

The Friedman test reports a significant overall difference among the four PPI models ($p = 7.8e\!-\!30$). The critical-difference diagram in Figure \ref{fig:critical_ppi} displays no connecting lines, indicating that every pair of methods differs by more than the critical difference. Our model achieves the highest average rank, with the Siamese Residual~RCNN in second place; the gap between them is statistically significant ($p = 1.5e\!-\!2$).

\begin{figure*}[t]
\centering
\includegraphics[width=\textwidth]{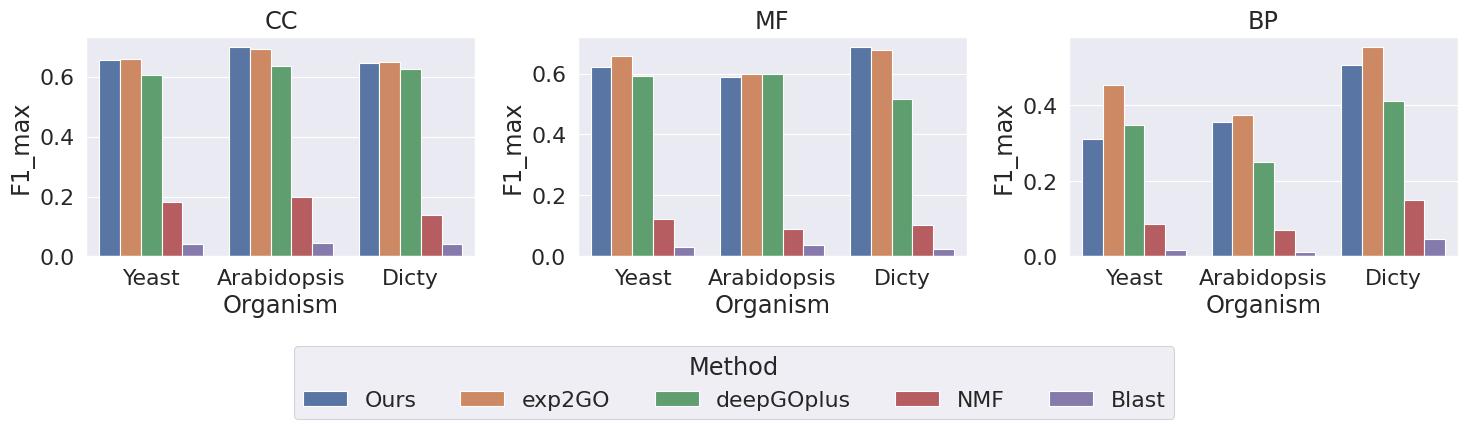}
\caption{Bar plot comparing the maximum F1 scores of different models across each subontology (Cellular Component, Molecular Function, and Biological Process) for each organism. The plot highlights the performance of our model against other methods (exp2GO, deepGOplus, and NMFGO) across all subontologies and organisms.}
\label{fig:bars}
\end{figure*}

\subsection{Gene Ontology (GO) term prediction}
For this task, we compared our method against a baseline sequence approach used in the CAFA challenge \cite{Zhou2019, Radivojac2013} (BLAST\cite{Altschul1990}) and three state-of-the-art approaches: NMF-GO \cite{Yu2020}, deepGOplus \cite{Kulmanov2020}, and exp2GO \cite{exp2go}. Following the experimental setup used by NMF-GO and exp2GO, we trained on historical GOA files (2016) and validated predictions using the 2017 GOA file.

The 2016 dataset was split into training and validation sets (16\% of the data) and the model was trained using the loss function described in Eq \ref{eq:loss} and the same tokenizer architecture (linear, ReLU, linear, Tanh), applied when node features were available, as in the case of PPI prediction.  The model was used to predict the semantic similarity matrix between genes, filling the gaps left because it was not possible to calculate semantic distances over the GO for unannotated genes. Early stopping with the validation set was performed by calculating the F1 score using a Bayesian probability method to predict the GO terms, as described in \cite{exp2go}. Five runs were performed with different initialization seeds, and the run with the lowest validation F1 score was selected. This approach efficiently evaluates the model's ability to predict GO terms on newly annotated proteins.

Figure \ref{fig:bars} presents a comparison of the different models across the datasets, organized into three subplots corresponding to the Molecular Function (MF), Biological Process (BP), and Cellular Component (CC) subontologies. Each subplot illustrates the performance of the models based on the $\text{F}1_{max}$ score across various species within the dataset, with the x-axis representing the different models and the y-axis indicating their respective $\text{F}1_{max}$ values. In the MF subontology, the performance among the models is closely matched, with the proposed model showing a slight improvement for the \textit{Dicty} species. Moving to the BP subontology, although the proposed model does not outperform exp2go, it still performs commendably, reflecting its capability in this area. Finally, in the CC subontology, the proposed model demonstrates a slight advantage over the others, highlighting its overall effectiveness in predicting functional similarity across diverse biological contexts.

To conduct a statistical analysis of the results, the Friedman test and critical difference diagram \cite{Friedman}, followed by the post-hoc Nemenyi test, were employed to assess the significance of the performance differences between the models.

\begin{figure}[h]
    \centering
    \includegraphics[width=\linewidth]{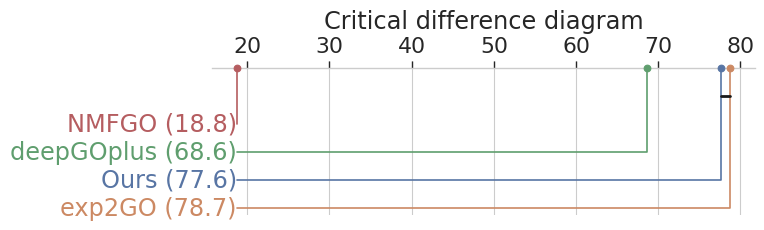}
    \caption{The critical difference (CD) diagram presents the statistical significance of the results. Models connected by a black line have no statistically significant difference. The CD value is 1.36 ($\alpha = 0.05$).}
    \label{fig:critical}
\end{figure}

The Friedman test results indicate that there are significant differences in the performance of the models being compared ($p=1e\!-\!6$). The critical difference diagram in Figure \ref{fig:critical} shows that both the proposed model and exp2GO are the best methods for gene function prediction, with no statistically significant difference between them ($p =0.237$). However, there is a significant difference between the proposed model and deepGOplus ($p =1e\!-\!3$) as well as NMFGO ($p =1e\!-\!3$).

\subsection{Compounds similarity prediction}
\begin{figure*}[!t]
\centering
\includegraphics[width=\textwidth]{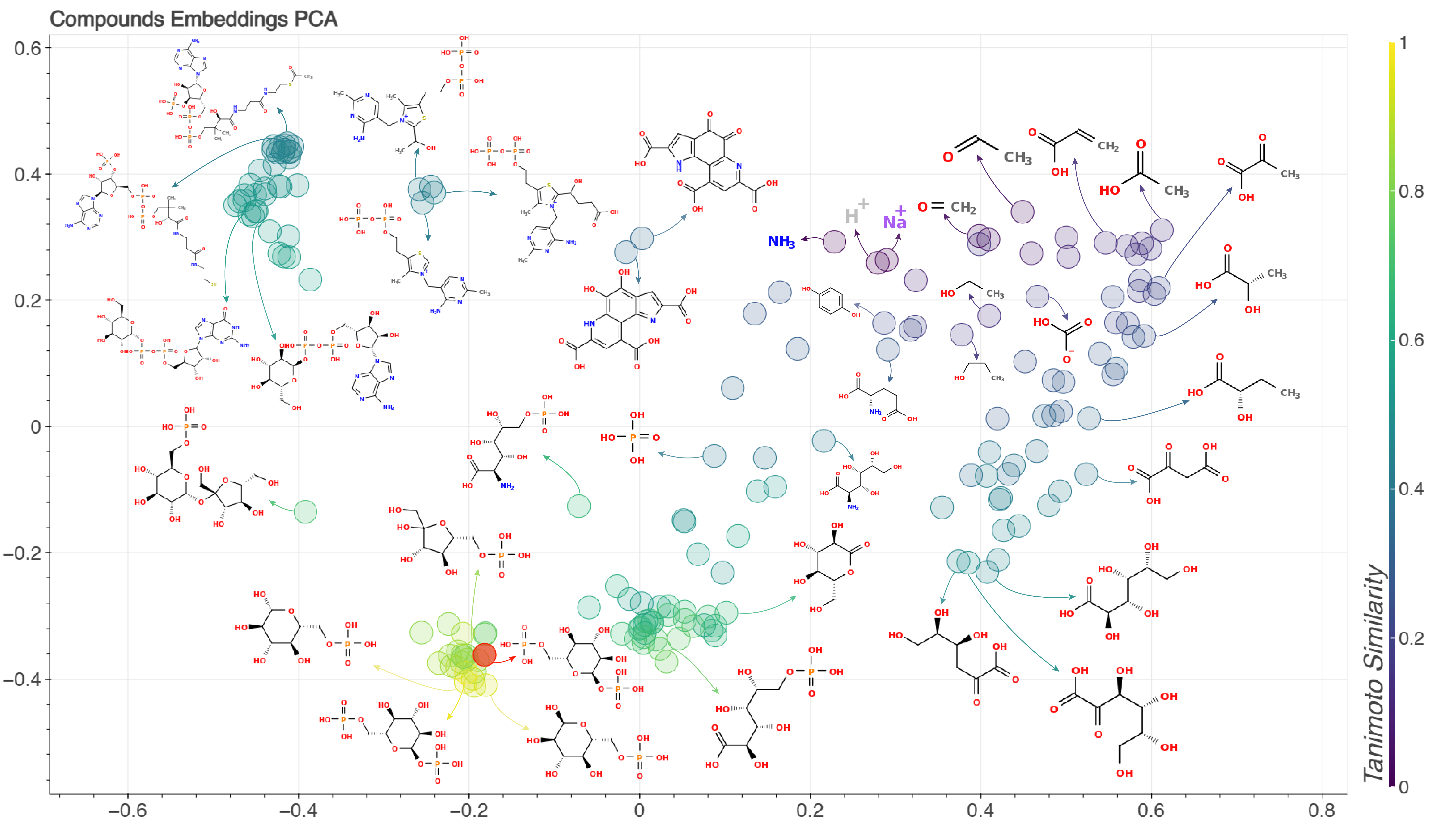}
\caption{PCA plot of embeddings for the analyzed compounds. Colors represent similarity to a reference compound (glucose-1-P), indicated in red. Similarity values decrease (colors shift towards violet) as the distance from the reference compound increases. Additionally, some compounds are shown with their molecular structures.}
\label{fig:pca_plot}
\end{figure*}

In this section, the findings on compound similarity are presented. While traditional approaches often rely on structural information, they face limitations when the compound structure is not available. We address this issue by using one-hot encoding as input features, enabling the model to predict similarity even in the absence of structural data.

For this task, the tokenizer employed a simpler architecture compared to previous experiments, consisting of a single linear layer followed by a Tanh activation function, effectively handling one-hot encoded features through extensive experimentation. Using this setup, we conducted a 5-fold cross-validation, obtaining a Mean Absolute Error (MAE) of 0.013. Given that the Tanimoto coefficient ranges from 0 to 1, this error represents only 1.3\%, indicating a high degree of accuracy in our model's predictions.

\begin{figure}[h]
    \centering
    \includegraphics[width=\linewidth]{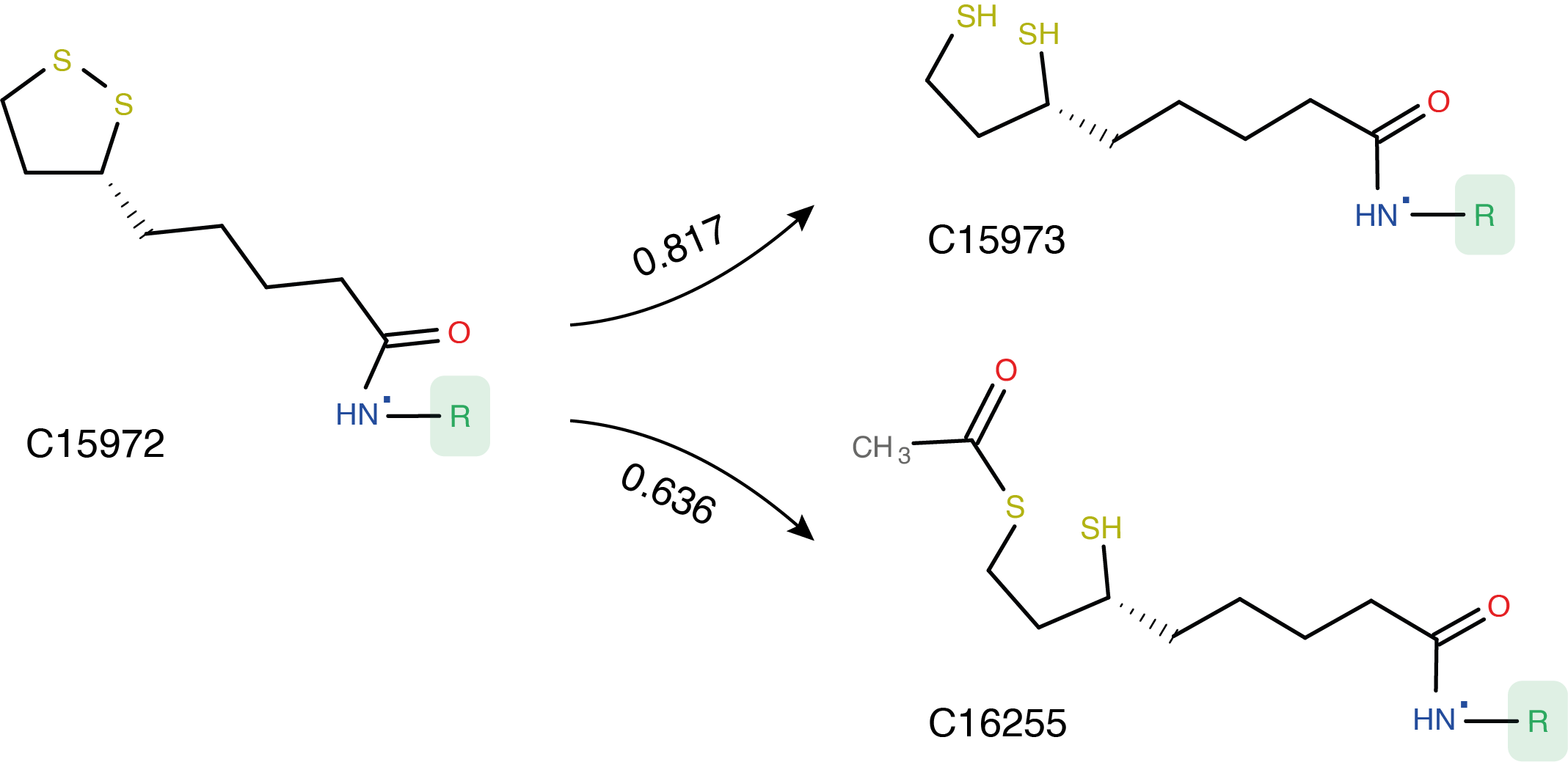}
    \caption{Partial structures of the compounds with KEGG ID C15972, C15973, and C16255. Generic substituents, marked in green as '-R', indicate that any functional group substituting for a hydrogen atom in the base compound structure, making it impossible to create a fingerprint for the compound. The arrows indicate predicted similarity values between compounds.}
    \label{fig:compoundsR}
    \vspace{-4pt}
\end{figure}

Figure \ref{fig:compoundsR} shows partial structures of the compounds with KEGG IDs C15972, C15973, and C16255. Although the molecular structures are partially known, each includes a generic substituent '-R', making it difficult to objectively assess similarity, as this information cannot be used to build fingerprints. This limitation, however, does not affect our approach, which instead represents compounds using embeddings learned from the graph topology of the metabolic pathway. For instance, our method predicts a similarity score of 0.817 between C15972 and C15973, which aligns with visual analysis as both compounds share a high proportion of their molecular structures. Similarly, a score of 0.636 between C15972 and C16255 reflects a moderate structural similarity, as these compounds share some structural elements but to a lesser extent than C15972 and C15973.

Figure~\ref{fig:pca_plot} presents a PCA projection plot of the embeddings for the analyzed compounds into the first two principal components, learned using our loss function. In this plot, colors indicate similarity to a reference compound, glucose-1-P, marked in red. As we move away from this reference compound, similarity values decrease gradually, with colors transitioning towards violet. This gradient effectively captures the similarity relationships within the dataset.
\vspace{-5pt}
\subsection*{Ablation study on compound similarity}

To better understand the individual contributions of the loss components and the attention mechanism, we performed an ablation study using the compound similarity dataset. Two independent experiments were carried out in addition to a basic \textit{baseline} MPNN. The attention type experiment kept the composite loss \(L_{\mathrm{s}}+L_{\mathrm{c}}+L_{\mathrm{cp}}\) fixed and evaluated four attention configurations: none, node only, edge only, and the combined node\&edge scheme. Conversely, the loss type experiment held the node\&edge attention constant and assessed four loss formulations: \(L_{\mathrm{s}}\), \(L_{\mathrm{s}}+L_{\mathrm{cp}}\), \(L_{\mathrm{s}}+L_{\mathrm{c}}\), and the full \(L_{\mathrm{s}}+L_{\mathrm{c}}+L_{\mathrm{cp}}\).

\begin{figure*}[!t]
    \centering
    \includegraphics[width=\textwidth]{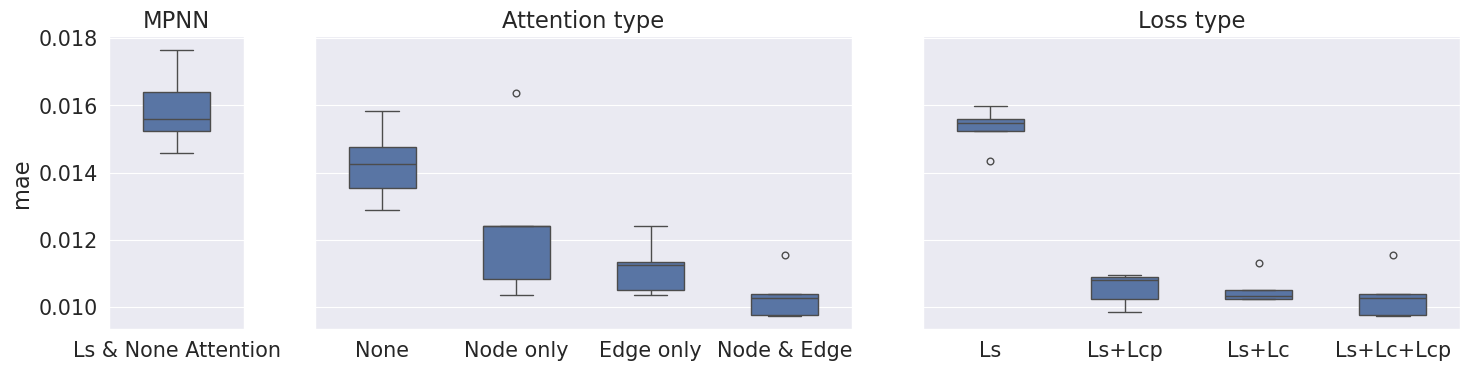}
    \caption{Mean Absolute Error (MAE) for the three models—baseline MPNN, attention experiment and loss experiment—presented in that order. Lower values indicate better performance.}
    \label{fig:boxplot_ablation}
\end{figure*}

\begin{figure}[!t]
    \centering
    \includegraphics[width=\linewidth]{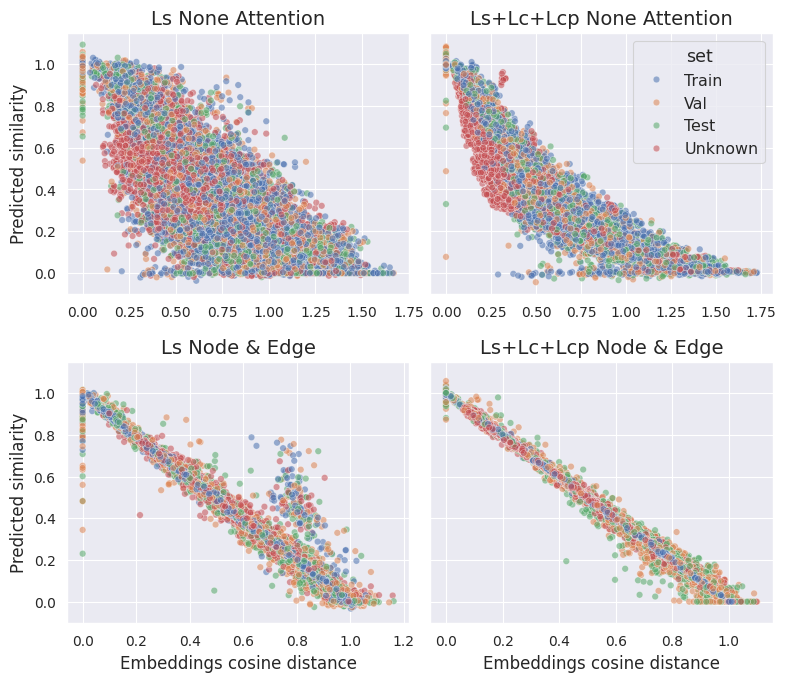}
    \caption{Predicted compound similarity (y-axis) versus cosine distance between embeddings (x-axis) under four ablation settings: baseline model without attention and self-supervised losses (top-left), attention only (top-right), hybrid loss only (bottom-left), and the full model with both attention and hybrid loss (bottom-right).}
    \label{fig:distplot_ablation}
\end{figure}

Figure~\ref{fig:boxplot_ablation} displays boxplots summarising the Mean Absolute Error obtained in each of the five folds of cross-validation for the baseline model and for every configuration evaluated in the attention-type and loss-type experiments. Each box therefore represents the distribution of errors across the 5-fold splits, providing a concise view of the variability and central tendency associated with each setting.

Figure \ref{fig:distplot_ablation} plots, for each compound pair, the cosine distance between embeddings on the horizontal axis and the similarity predicted by the model on the vertical axis. The four panels correspond to the baseline configuration, the model with attention only, the model with the hybrid loss only, and the full model. In the baseline panel, the points form a wide, diffuse cloud, indicating that embedding distance is not a good proxy for similarity prediction. Enabling either component separately narrows the cloud and makes the trend more clearly monotonic, showing that each element independently helps the embeddings capture similarity. When both the attention mechanism and the hybrid losses are active, the points collapse into an almost straight, tight band; the markedly lower dispersion indicates that the embeddings are better organised in the latent space, thereby enabling the model to predict similarity much more accurately. This improved organisation not only accounts for the lower MAE observed but also ensures that compounds lacking structural information are embedded coherently alongside known compounds, providing a reliable latent representation for downstream analyses.

\section{Conclusion}
\label{sec:conclusion}
This study introduces a novel model based on Message Passing Neural Networks (MPNNs) designed to address edge-centric tasks in graph-based learning. By integrating an attention mechanism, the proposed architecture effectively utilizes both node features and edge attributes, enhancing its performance in edge regression and classification tasks where traditional methods have often faced limitations.

The custom loss function introduced in this model combines supervised and self-supervised learning, allowing it to optimize both predictions and embeddings simultaneously. This approach ensures robust generalization, even in scenarios with limited or missing information about nodes. By utilizing cosine similarity between node embeddings and predictions, the model effectively organizes learned representations, facilitating the modeling of complex relationships within dynamic graphs.

Experimental results across multiple datasets demonstrate that the proposed model outperforms state-of-the-art methods in protein-protein interaction (PPI) prediction while remaining competitive in Gene Ontology (GO) term prediction. The model also demonstrates that k-Nearest Neighbors Graphs (KNNG) based on node similarity measures are effective for graph construction, enhancing the model’s capacity to propagate information from neighboring nodes and enrich node descriptors.

Moreover, the model presents a novel solution to the problem of predicting similarity between compounds with unknown structure, using one-hot encoding as node features to compensate for the absence of detailed structural information. This approach not only enables accurate similarity predictions without requiring compound structure but also proves valuable in scenarios with limited or missing node features, generating meaningful representations even with sparse data. Its effectiveness in this area suggests potential applications in drug discovery, molecular similarity assessment, and similarity search.

In conclusion, the proposed model offers a flexible and powerful solution for a wide range of graph-based applications, particularly for tasks that require evaluating the properties of pairs of objects. Future research may focus on extending its capabilities to other fields of application, for example in recommendation systems, and social network analytics, among others that may benefit from the capabilities of predicting properties of pairs of nodes. In all our experiments we have used a one-layer MPNN, but we need to explore the possibility of using additional layers of MPNN in the Embedder section. Furthermore, future investigations will consider methods to generate predictions with associated uncertainty intervals, providing a measure of confidence in the model predictions.

\bibliographystyle{IEEEtran}
\bibliography{bibliography}

\end{document}